# Boosting Adversarial Transferability Against Defenses via Multi-Scale Transformation

Zihong Guo, Chen Wan[✉], Yayin Zheng, Hailing Kuang, Xiaohai Lu

Department of Computer Science and Technology, Shantou University, Shantou, China
`wanchen18@outlook.com`

**Abstract.** The transferability of adversarial examples poses a significant security challenge for deep neural networks, which can be attacked without knowing anything about them. In this paper, we propose a new Segmented Gaussian Pyramid (SGP) attack method to enhance the transferability, particularly against defense models. Unlike existing methods that generally focus on single-scale images, our approach employs Gaussian filtering and three types of downsampling to construct a series of multi-scale examples. Then, the gradients of the loss function with respect to each scale are computed, and their average is used to determine the adversarial perturbations. The proposed SGP can be considered an input transformation with high extensibility that is easily integrated into most existing adversarial attacks. Extensive experiments demonstrate that in contrast to the state-of-the-art methods, SGP significantly enhances attack success rates against black-box defense models, with average attack success rates increasing by 2.3% to 32.6%, based only on transferability.

**Keywords:** Adversarial examples, Transferability, Multi-scale transformation.

## 1 Introduction

Deep neural networks (DNNs) have achieved remarkable success across a wide range of applications, such as image classification [1], autonomous driving [2], and face recognition [3]. However, despite their impressive performance, DNNs are vulnerable to adversarial examples, which are crafted by adding small perturbations to the original image and can mislead the model into making incorrect predictions [4,5]. Moreover, the adversarial examples generally exhibit transferability, meaning that adversarial examples generated on one model can effectively mislead another model, even if the latter has a different architecture or parameters (i.e., black-box attacks). This property makes transfer-based attacks particularly applicable to real-world models, and the study of transferability has attracted significant attention.

To enhance the transferability of adversarial examples, various techniques have been proposed, including advanced gradient computations [6], ensemble-model attacks [7], input transformations [8,9], and model-specific methods [10]. Among them, input transformation has emerged as a popular strategy, which aims to prevent adversarial examples from overfitting to the white-box model. A series of transformation strategies have been proposed, e.g., the Diverse Input Method (DIM) [11], Translation-Invariant

Method (TIM) [12], Scale-Invariant Method (SIM) [13], Admix [8], and BSR [9]. These methods transform the input image to obtain the augmented example and then feed it into the model for gradient computation, which is used to determine the adversarial perturbations.

Most of the existing input transformations are mainly focused on single-scale images and have not yet taken multi-scale features (i.e., different resolutions of the same image) into account. This limitation prompts us to consider whether the transferability can be further enhanced by introducing a multi-scale processing mechanism of the human visual system. When humans observe a complex scene, the visual system processes and integrates information across various scales, combining both global structures and fine-grained details for recognition. This multi-scale processing ability allows humans to accurately recognize complex environments.

In this paper, we explore the multi-scale based adversarial attack and propose a new approach called Segmented Gaussian Pyramid (SGP). To illustrate our method, Fig. 1 provides an example of the three-layer SGP. The initial input, depicted in Fig. 1(a), is first convolved with a Gaussian kernel and then subjected to three distinct sampling schemes, i.e., row&column, row-only, and column-only, to yield the three different versions in the second layer (shown Fig. 1(b)–Fig. 1(d)). Subsequently, the image generated through row&column sampling (i.e., Fig. 1(b)) is transferred to the third layer and undergoes the same convolution and sampling operations as before. The resulting outputs are illustrated in Fig. 1(e)–Fig. 1(g). As observed, the three-layer SGP yields seven multi-scale examples, i.e., one original and six augmented versions. After obtaining these multi-scale examples, we calculate the gradient of the loss function with respect to each scale and average the results to determine the adversarial perturbations. Our approach leverages multi-scale examples to comprehensively capture both global and local features of the input image, which can enhance the effectiveness of adversarial attacks.

The main contributions can be summarized as follows:

- In this paper, we explore the multi-scale based adversarial attack and propose a new Segmented Gaussian Pyramid (SGP) attack method.
- Extensive experiments demonstrate that our approach achieves higher attack success rates in black-box defense models, and the generated adversarial examples exhibit excellent transferability.
- The proposed SGP can be viewed as an input transformation with high extensibility that seamlessly integrates into most existing adversarial attacks. For instance, introducing our method into BSR [9] improves the average attack success rate by 33.7% across seven black-box defense models.

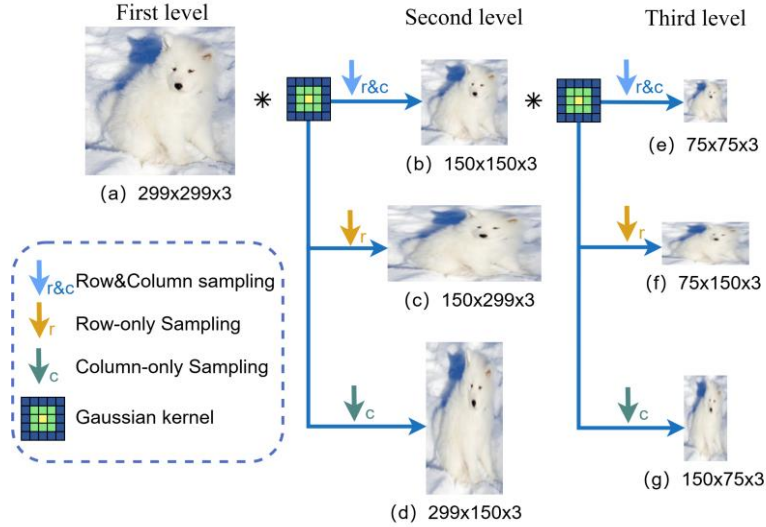

**Fig. 1.** The flowchart of our proposed method, wherein * symbolizes the convolution operation.

## 2  Related Work

According to the level of knowledge about the model, existing attack algorithms can be classified into two main categories, i.e., white-box and black-box. In white-box attacks, all the information about the target model can be accessed, including the architecture and weights. However, in the real world, it is generally difficult for the adversary to obtain all this information, making black-box attacks more practical. The exploitation of the transferability of adversarial examples is a common method in black-box attacks. In this paper, we mainly focus on black-box attacks against defense models using the transferability of adversarial examples, which can further validate the robustness of the defense approach.

Recently, many methods have been proposed to enhance the transferability of adversarial examples. Among these, advanced gradient calculations focus on improving optimization algorithms (i.e., by introducing the advanced gradient terms) to prevent adversarial examples from falling into poor local maxima. For example, MI-FGSM [14] introduces the momentum term, and NI-FGSM [13] employs Nesterov's accelerated gradient term. In addition, some input transformation techniques have been proposed to avoid the generated adversarial examples overfitting the model. Diverse Input Method (DIM) [11] randomly resizes and pads the input image with a fixed probability. Translation-Invariant Method (TIM) [12] transforms an image into a set of images for gradient computation, approximated by convolving the gradient with a Gaussian kernel. Scale-Invariant Method (SIM) [13] computes the gradient of multiple scaled images. Admix [8] mixes a small set of images from other categories into the input example to form a set of mixed images. BSR [9] divides the input image into blocks, randomly shuffles, and rotates them to construct a series of augmented examples.

## 3   Methodology

### 3.1   Motivation

Adversarial attacks in existing research generally focus on single-scale features, with limited attention given to the integration of global and local information across different scales. Drawing inspiration from the human visual system that excels at integrating multi-scale information, we propose a new segmented Gaussian pyramid (SGP) attack method. The proposed method leverages the multi-scale properties of input examples to enhance the generalization of perturbations, making them more effective across diverse models and scenarios. As pointed out in the previous study [15], a deeper comprehension of both global and local features in input examples significantly improves the cross-model transferability of adversarial examples, which serves as the core motivation behind our method.

As mentioned earlier, a total of seven examples can be constructed via the three-layer SGP, including one original example and six multi-scale examples, as shown in Fig. 1(a)-Fig. 1(g). We show the attention heatmaps of the Inc-v3 model for these examples in Fig. 2, to represent the discriminative regions used for predictions. Please note that to ensure compatibility when feeding the six multi-scale examples into the model, we resize them to match the dimensions of the original example. Fig. 2(a) displays the heatmap corresponding to the initial input (i.e., Fig. 1(a)). Fig. 2(b)-Fig. 2(d) represent the heatmaps for the examples depicted in the second layer (i.e., Fig. 1(b)-Fig. 1(d)), and Fig. 2(e)-Fig. 2(g) illustrate the heatmaps for the examples in the third layer (i.e., Fig. 1(e)-Fig. 1(g)), respectively. As seen, the salient region undergoes a slight displacement, suggesting that the model adaptively reorients its focus when predicting multi-scale examples. Meanwhile, with an increasing number of pyramid layers, the regions covered by the attention heatmap progressively expand, meaning the model can capture more global and local features across multi-scale examples. This improvement enhances the capacity of the model to understand complex data, which can be leveraged to develop more effective adversarial perturbations.

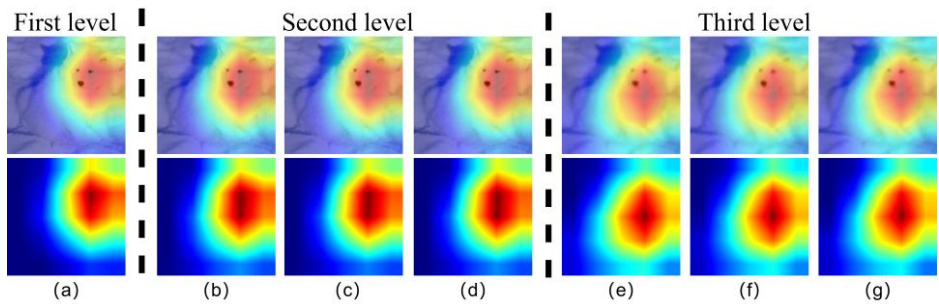

**Fig. 2.** Attention heatmaps of SGP-Generated images on Inc-v3 using Grad-CAM. Segmented Gaussian Pyramid.

### 3.2    Segmented Gaussion Pyramid

Let $x$ represent an original image and $y$ denote its corresponding ground-truth label. The network parameters are denoted by $\theta$, and the loss function is represented as $J(x; y; \theta)$. The objective of generating an adversarial example $x^{adv}$ is to maximize the loss function $J(x + r; y; \theta)$, subject to the constraint that the adversarial example $x^{adv} = x + r$ remains visually similar to the original image $x$ while ensuring that the predicted label $y^{pre} \neq y$. In this work, we utilize the $l_\infty$-norm to quantify the perceptibility of the adversarial perturbations, constrained by $\left\|r\right\|_\infty \leq \varepsilon$

The essence of our proposed SGP is the construction of a Gaussian pyramid that augments the input example across various scales. This approach diversifies the input data by generating multi-scale examples, which can be viewed as an input transformation strategy. In our SGP, each layer involves two fundamental operations, namely, filtering and sampling. The image is first smoothed with a Gaussian kernel, and then three multi-scale examples are generated through various sampling schemes, i.e., row&column sampling, row-only sampling, and column-only sampling. Different from the traditional Gaussian pyramid that only generates a single-resolution image per layer, our method produces three different resolutions at each layer. The diversity of input example can be further enhanced, which allows the model to capture a wider range of features at different scales.

Without loss of generality, we assume that the current iteration is the $t$-th iteration, the input example is $x_t^{adv}$, and the crafted adversarial example is $x_{t+1}^{adv}$. The proposed SGP performs the multi-scale transformation operation on the input example $x_t^{adv}$ to construct a series of multi-scale examples. In the filtering stage, we first smooth the example $x[i]$ (where $x[i]$ is defined below) using a Gaussian kernel $K$ to reduce high-frequency noise and enhance the relevant features. This process is carried out as follows,

$$K = \frac{1}{256} * \begin{bmatrix} 1 & 4 & 6 & 4 & 1 \\ 4 & 16 & 24 & 16 & 4 \\ 6 & 24 & 36 & 24 & 6 \\ 4 & 16 & 24 & 16 & 4 \\ 1 & 4 & 6 & 4 & 1 \end{bmatrix} \tag{1}$$

$$x[i+1] = x[i] * K \tag{2}$$

where $*$ denotes the convolution operation, $i$ denotes the $i$-th layer in SGP, and $m$ is a hyper-parameter that represents the total number of pyramid layers. When $i = 1$, $x[1]$ is the input example $x_t^{adv}$ at the $t$-th iteration, i.e., $x[1] = x_t^{adv}$; when $i > 1$, $x[i]$ is the image obtained by row&column sampling of the previous layer $x[i-1]$ (i.e., $x[i-1] \downarrow_{r\&c}$, please refer to the sampling operation below). For example, at the third layer, we perform Gaussian filtering on the image obtained through row&column sampling (i.e., Fig. 1(b)).

In the sampling phase, three downsampling schemes, namely row&column sampling $\downarrow_{r\&c}$, row-only sampling $\downarrow_r$, and column-only sampling $\downarrow_c$, are chosen to perform odd row and odd column sampling operations on the Gaussian-filtered example $x[i+1]$, as follows,

$$\mathcal{T}[i+1,j] = \begin{cases} x[i+1]\downarrow_{r\&c} & if\ j=1 \\ x[i+1]\downarrow_r & if\ j=2 \\ x[i+1]\downarrow_c & if\ j=3 \end{cases} \quad (3)$$

where $j=1,2$, and $3$ denote row&column sampling $\downarrow_{r\&c}$, row-only sampling $\downarrow_r$, and column-only sampling $\downarrow_c$, respectively. $\mathcal{T}[i+1,j]$ denotes the set of augmented examples obtained by sampling $x[i+1]$ using the $j$-th scheme for the $(i+1)$-th pyramid layer, $0 \le i \le m-1$, and $m$ represents the total number of pyramid layers. When $m=1$, there is only one example, namely the input example $x_t^{adv}$. When $m>1$, three augmented examples are constructed at each pyramid layer, and there are $3m-2$ multi-scale examples in total, including one input example and $3(m-1)$ augmented examples.

### 3.3  Attack Method

In this part, we illustrate how multi-scale examples can be utilized to determine the perturbations added to the current example $x_t^{adv}$. To ensure model compatibility, the dimensions of the multi-scale examples are adjusted same as the original image. Then, we feed the multi-scale examples into the model and compute the gradient of the loss function with respect to each scale separately. These individual gradients are averaged to derive a composite gradient, as fallows,

$$\bar{g}_t = \begin{cases} \nabla_{x_t^{adv}} J(x_t^{adv}; y; \theta) & if\ m=1 \\ \dfrac{1}{3m-2}\left(\nabla_{x_t^{adv}} J(x_t^{adv}; y; \theta) + G_t\right) & if\ m>1 \end{cases} \quad (4)$$

where $\nabla_{x_t^{adv}} J(x_t^{adv}; y; \theta)$ represents the gradient of the loss function with respect to the input example $x_t^{adv}$, $G_t = \sum_{i=2}^{m}\sum_{j=1}^{3}\nabla_{x_t^{adv}} J(R(\mathcal{T}[i,j]); y; \theta)$ denotes the accumulation of the gradient of the loss function with respect to all augmented examples, and $R(\cdot)$ denotes the resize operation. According to the composite gradient $\bar{g}_t$ obtained in Eq. (4), we can determine the adversarial perturbations that are added to the current example $x_t^{adv}$ to craft the adversarial example $x_{t+1}^{adv}$.

In practice, the composite gradient $\bar{g}_t$ can be readily introduced into most existing adversarial attacks to form a series of advanced methods. We summarize the algorithm of SGP integrated into MI-FGSM (denoted as SGP without ambiguity in the following) in Algorithm 1. If the number of pyramid layers $m=1$, SGP degenerates to MI-FGSM. In our proposed SGP, we need to calculate the gradient of the loss function with respect to $(3m-2)$ multi-scale examples at each iteration. Thus, the number of gradient calculations for generating the adversarial examples $x_t^{adv}$ is $(3m-2)*T$ in total. Since multi-scale examples are used to determine the added perturbation in SGP, the transferability can be significantly improved, as demonstrated by our extensive experimental results.

---

**Algorithm 1** Segmented Gaussian Pyramid

---

**Input:** A classifier $f$ with loss function $J$; a benign example $x$ with ground-truth label $y$; the maximum perturbation $\epsilon$; number of iterations $T$; decay factor $\mu$; number of pyramid layers $m$

**Output:** An adversarial example $x^{adv}$

1: Let $\alpha = \frac{\epsilon}{T}$; $g_0 = 0$; $\bar{g}_0 = 0$; $x_0^{adv} = x$
2: for $t = 0$ to $T - 1$ **do**
3:   Construct the multi-scale examples by Eq. (3)
4:   Calculate the composite gradient $\bar{g}_t$ by Eq. (4)
5:   Update the enhanced momentum $g_{t+1}$:
$$g_{t+1} = \mu \cdot g_t + \frac{\bar{g}_t}{|\bar{g}_t|_1}$$
6:   Update $x_{t+1}^{adv}$ by applying the gradient sign:
$$x_{t+1}^{adv} = x_t^{adv} + \alpha \cdot sign(g_{t+1})$$
7: **end for**
8: **return** $x^{adv} = x_T^{adv}$

---

## 4 EXPERIMENTS

### 4.1 Experimental Setup

Our method is implemented using TensorFlow and evaluated primarily on an NVIDIA RTX 4090 Ti GPU.

**Dataset.** The test dataset consists of 1000 randomly selected images from the ImageNet validation set provided by Lin *et al.* [13].

**Models.** In our experiments, seven defense models are selected for testing, including Inc-v3$_{ens3}$, Inc-v3$_{ens4}$, IncRes-v2$_{ens}$ [16], HGD [17], R&P [18], NIPS-r3[1], and NPR [19]. To attack these defenses based on the transferability, we choose four normally trained models, *i.e.*, Inc-v3 (Iv3) [21], Inc-v4 (Iv4), IncRes-v2 (IRv2) [22], and Res-101 (R101) [23], as the white-box models to craft adversarial examples.

**Baseline.** five competitive input transformations, *i.e.*, DIM [11], TIM [12], SIM [13], Admix [8] and SSA [24], along with their combined variant (denoted as STDM) serve as baselines for comparison with our proposed SPG. Furthermore, to demonstrate the extensibility of our approach, we introduce SGP into six baselines (i.e., DIM [11], TIM [12], SIM [13], Admix [8], SSA [24], and BSR [9]) for comparative evaluation. All input transformations are integrated into MI-FGSM [14].

**Hyper-parameters.** The maximum perturbation, number of iterations, step size, and decay factor are set to $\epsilon = 16, T = 10, \alpha = \epsilon/T = 1.6, \mu = 1$. For the five data augmentation strategies, the transformation probability of DIM is set to 0.5, the kernel matrix of TIM is Gaussian kernel with the size of $7 \times 7$, the number of scale copies in SIM is set to 5, in Admix 3 images with the intensity of 0.2 are admixed, the turning factor in SSA was set to 0.5 and the standard deviation was set to $\epsilon$. and in BSR image split into 2 × 2 blocks with a maximum rotation angle of 24°. The number of layers $m$ in our SGP is set to 3.

---

[1] https://github.com/anlthms/nips-2017/tree/master/mmd

### 4.2  Evaluation on Single Input Transformation

We first evaluate the attack success rate of various single input transformations, namely DIM, TIM, SIM, Admix, SSA and our proposed SGP. The attack success rates of the obtained adversarial examples against the seven defense models are shown in Table 1. As seen, in contrast to baselines, the attack success rate of SPG improved by an average of 2.3% to 32.6%. For instance, the average success rate of SPG-induced attacks is improved by 19.8% (DIM), 15.8% (TIM), 9.9% (SIM), 4.3(Admix) and 3.7% (SSA) compared to the three baselines for adversaries crafted on the Inc-v3 model. The experimental results demonstrate that the adversarial examples generated by our proposed SGP exhibit greater aggressiveness across various defense models. This improvement may be attributed to the introduction of multi-scale processing, which enhances the deep understanding of the input image by the model.

**Table 1.** Attack success rate (%) of the seven defense models under a single-model setup with various single input transformations. The adversarial examples are generated via Iv3(Inc-v3), Iv4(Inc-v4), IRv2(IncRes-v2), and R101(Res-101), respectively.

| Model | Attack | Inc-v3$_{ens3}$ | Inc-v3$_{ens4}$ | IncRes-v2$_{ens}$ | HGD | R&P | NIPS-r3 | NPR | Avg. |
|---|---|---|---|---|---|---|---|---|---|
| Iv3 | DIM | 18.5 | 17.6 | 9.4 | 6.9 | 8.0 | 14.2 | 16.1 | 13.0 |
|  | TIM | 24.1 | 21.0 | 12.8 | 16.9 | 12.0 | 15.1 | 17.3 | 17.0 |
|  | SIM | 32.6 | 31.1 | 17.6 | 14.7 | 16.4 | 23.7 | 24.3 | 22.9 |
|  | Admix | 40.1 | 39.0 | 21.1 | 21.4 | 20.6 | 28.9 | 28.1 | 28.5 |
|  | SSA | 40.6 | 38.2 | 23.3 | 14.5 | 23.0 | 30.2 | 33.9 | 29.1 |
|  | SGP (Ours) | **42.2** | **42.6** | **27.2** | **28.8** | **25.3** | **31.8** | **31.9** | **32.8** |
| Iv4 | DIM | 22.1 | 20.8 | 10.1 | 10.6 | 12.3 | 16.7 | 15.5 | 15.4 |
|  | TIM | 25.9 | 23.9 | 17.5 | 22.5 | 17.1 | 19.4 | 15.9 | 20.3 |
|  | SIM | 47.5 | 45.1 | 29.3 | 29.4 | 29.2 | 36.6 | 29.8 | 35.3 |
|  | Admix | **50.9** | **51.4** | 33.0 | 35.9 | 33.8 | 40.1 | 33.9 | 39.9 |
|  | SSA | 46.1 | 43.8 | 31.9 | 25.9 | 29.2 | 39.1 | **36.3** | 36.0 |
|  | SGP (Ours) | 49.9 | 50.2 | **37.3** | **39.4** | **34.7** | **40.2** | 35.8 | **41.1** |
| IRv2 | DIM | 32.2 | 25.1 | 17.7 | 19.2 | 18.3 | 23.8 | 19.2 | 22.2 |
|  | TIM | 32.2 | 27.7 | 21.9 | 26.2 | 21.0 | 24.0 | 19.7 | 24.7 |
|  | SIM | 57.1 | 49.1 | 41.7 | 40.0 | 35.8 | 43.1 | 34.5 | 43.0 |
|  | Admix | 61.1 | 52.1 | 45.7 | 43.5 | 40.1 | 48.6 | 39.4 | 47.2 |
|  | SSA | 56.5 | 52.1 | 46.1 | 42.5 | 42.3 | 49.5 | 42.4 | 47.3 |
|  | SGP (Ours) | **61.7** | **59.9** | **56.0** | **53.1** | **50.1** | **54.4** | **48.3** | **54.8** |
| R101 | DIM | 36.1 | 32.7 | 20.9 | 27.8 | 22.4 | 30.3 | 24.4 | 27.8 |
|  | TIM | 36.4 | 32.3 | 22.9 | 31.4 | 24.1 | 28.2 | 24.3 | 28.5 |
|  | SIM | 43.4 | 38.3 | 26.3 | 32.3 | 25.8 | 33.0 | 29.0 | 32.6 |
|  | Admix | 48.7 | 42.1 | 30.5 | 35.8 | 29.0 | 37.9 | 32.3 | 36.6 |
|  | SSA | 53.0 | 50.2 | 39.1 | 44.3 | 37.9 | 47.3 | 41.7 | 44.8 |
|  | SGP (Ours) | **56.7** | **54.4** | **42.8** | **46.2** | **39.6** | **47.4** | **42.9** | **47.1** |

### 4.3 Evaluation on Combined Input Transformation

To demonstrate the extensibility of the proposed SGP, we introduce SGP into seven baseline methods (i.e., TIM, DIM, SIM, STDM, Admix, SSA, and BSR) that can form seven SGP-based attacks, named SGP-TIM, SGP-DIM, SGP-SIM, SGP-SI-TI-DM, SGP-Admix, SGP-SSA, and SGP-BSR. These SGP-based attacks and their corresponding baseline methods are used to attack Inc-v3 to generate adversarial examples, and then the attack success rate of these adversarial examples against the above-mentioned seven defense models is shown in Table 2. As seen, the average attack success rate of our proposed SGP-based methods is generally improved by a range of 10.6% to 33.7% compared with the corresponding baseline methods on Inc-v3. These results demonstrate that the proposed SGP can be viewed as an input transformation with high extensibility, which can be seamlessly integrated into most existing adversarial attacks to enhance their transferability.

**Table 2.** Attack success rate (%) of the seven defense models under a single-model setup with various single input transformations combined with SGP. The adversarial examples are generated via Inc-v3. ↑ represents the increase in attack success rate after being combined with SGP.

| Attack | Inc-v3$_{ens3}$ | Inc-v3$_{ens4}$ | IncRes-v2$_{ens}$ | HGD | R&P | NIPS-r3 | NPR | Avg. |
|---|---|---|---|---|---|---|---|---|
| SGP-TIM | 48.5$_{\uparrow 24.4}$ | 48.0$_{\uparrow 27.0}$ | 33.2$_{\uparrow 20.4}$ | 33.6$_{\uparrow 16.7}$ | 30.3$_{\uparrow 18.3}$ | 35.0$_{\uparrow 19.9}$ | 35.7$_{\uparrow 18.4}$ | 37.8$_{\uparrow 20.8}$ |
| SGP-DIM | 56.7$_{\uparrow 38.2}$ | 58.2$_{\uparrow 40.6}$ | 40.1$_{\uparrow 30.7}$ | 42.2$_{\uparrow 35.3}$ | 38.2$_{\uparrow 30.2}$ | 45.4$_{\uparrow 31.2}$ | 41.0$_{\uparrow 24.9}$ | 46.0$_{\uparrow 33.0}$ |
| SGP-SIM | 62.2$_{\uparrow 29.6}$ | 64.2$_{\uparrow 33.1}$ | 45.7$_{\uparrow 28.1}$ | 48.4$_{\uparrow 33.7}$ | 40.6$_{\uparrow 24.4}$ | 49.8$_{\uparrow 26.1}$ | 46.1$_{\uparrow 21.8}$ | 51.0$_{\uparrow 28.1}$ |
| SGP-STDM | 74.5$_{\uparrow 9.3}$ | 73.6$_{\uparrow 10.3}$ | 60.4$_{\uparrow 13.9}$ | 59.9$_{\uparrow 2.7}$ | 56.4$_{\uparrow 9.7}$ | 62.6$_{\uparrow 7.8}$ | 61.1$_{\uparrow 20.1}$ | 64.1$_{\uparrow 10.6}$ |
| SGP-Admix | 66.0$_{\uparrow 32.7}$ | 68.7$_{\uparrow 37.0}$ | 50.2$_{\uparrow 33.6}$ | 53.7$_{\uparrow 37.7}$ | 47.0$_{\uparrow 30.4}$ | 55.4$_{\uparrow 32.3}$ | 50.0$_{\uparrow 25.8}$ | 55.9$_{\uparrow 27.4}$ |
| SGP-SSA | 71.5$_{\uparrow 30.9}$ | 71.6$_{\uparrow 33.4}$ | 56.8$_{\uparrow 33.5}$ | 57.3$_{\uparrow 42.8}$ | 52.2$_{\uparrow 29.2}$ | 60.7$_{\uparrow 30.5}$ | 59.4$_{\uparrow 25.5}$ | 61.4$_{\uparrow 32.3}$ |
| SGP-BSR | 84.9$_{\uparrow 31.1}$ | 83.7$_{\uparrow 31.1}$ | 67.3$_{\uparrow 36.5}$ | 77.0$_{\uparrow 34.7}$ | 68.7$_{\uparrow 35.7}$ | 76.1$_{\uparrow 32.3}$ | 61.0$_{\uparrow 32.2}$ | 74.1$_{\uparrow 33.7}$ |

### 4.4 Evaluation of Attacks on the Ensemble Models

Liu *et al.* [16] showed that attacking multiple models simultaneously improves the transferability of the generated adversarial examples. To further demonstrate the effectiveness of the proposed SGP, we introduced the SGP in three baseline methods (i.e., SI-TI-DIM, Admix, SSA, and BSR), and these attacks and their corresponding baseline methods are used to attack the ensemble models, which consist of Inc-v3 and Inc-v4 with the same weights. Table 3 shows the success rates of these obtained adversarial examples for attacking the above seven defense models. As seen, the average attack success rate of our methods is about 12.3% to 35.6% higher than that of the baseline methods.

**Table 3.** Attack success rate (%) of seven defense models in the setting of ensemble models with various input transformations combined with SGP. The ensemble models consist of Inc-v3 and Inc-v4. ↑ represents the increase in attack success rate after being combined with SGP.

| Attack | Inc-v3$_{ens3}$ | Inc-v3$_{ens4}$ | IncRes-v2$_{ens}$ | HGD | R&P | NIPS-r3 | NRP | Avg. |
|---|---|---|---|---|---|---|---|---|
| SGP-STDM | 89.5$_{↑9.5}$ | 90.0$_{↑11.5}$ | 82.5$_{↑13.2}$ | 85.1$_{↑7.8}$ | 80.6$_{↑11.3}$ | 84.1$_{↑10.7}$ | 78.1$_{↑21.8}$ | 84.3$_{↑12.3}$ |
| SGP-Admix | 87.0$_{↑27.4}$ | 87.3$_{↑32.0}$ | 77.3$_{↑42.3}$ | 80.3$_{↑38.9}$ | 73.2$_{↑35.8}$ | 80.4$_{↑33.0}$ | 72.6$_{↑39.9}$ | 79.7$_{↑35.6}$ |
| SGP-SSA | 85.6$_{↑26.1}$ | 85.6$_{↑27.5}$ | 75.4$_{↑37.1}$ | 77.7$_{↑45.4}$ | 71.9$_{↑33.1}$ | 79.5$_{↑28.6}$ | 76.5$_{↑29.0}$ | 85.6$_{↑26.1}$ |
| SGP-BSR | 92.5$_{↑19.3}$ | 92.2$_{↑24.1}$ | 83.5$_{↑38.3}$ | 90.8$_{↑21.9}$ | 84.9$_{↑32.7}$ | 88.2$_{↑21.6}$ | 73.0$_{↑34.7}$ | 86.4$_{↑27.5}$ |

### 4.5 Ablation Study

The proposed SGP introduces a new hyper-parameter, the number of pyramid layers $m$. In order to study the effect of $m$ on the attack success rate, we use SGP to attack Inc-v3 to generate adversarial examples, where the number of pyramid layers $m$ ranges from 1 to 8. The obtained adversarial examples are tested on three defense models (i.e., Inc-v3$_{ens3}$, Inc-v3$_{ens4}$, and IncRes-v2$_{ens}$), and the experimental results are shown in Fig. 3. As observed, when the number of pyramid layers is approximately 3, our proposed SGP generally can achieve the highest attack success rate. This is because, when the number of pyramid layers $m \leq 3$, the ability of multi-scale feature extraction enables the model to flexibly capture rich local and global information, which significantly improves the attack success rate; however, once the number of layers $m > 3$, the gain of feature extraction tends to be saturated, resulting in a stabilization of the attack success rate. Considering that the larger $m$ would increase the computation time, it is suggested to set $m$ to 3 in practice for a balanced trade-off between efficiency and performance.

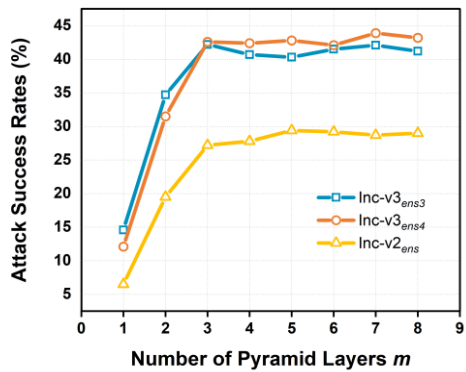

**Fig. 3.** The impact of the number of pyramid layers $m$ on the attack success rate (%).

## 5   Conclusion

In this study, we propose a new SGP attack method that leverages the multi-scale features of input examples to enhance the transferability of adversarial examples. As an input transformation strategy, our proposed method has high extensibility and can be easily integrated into most existing adversarial attacks. Extensive experimental results demonstrate that based only on transferability, SGP can significantly improve the attack success rate against black-box defense models. In future work, we plan to further investigate the multi-scale features of images and explore their potential in generating more effective adversarial examples.

**Acknowledgments.** This research is funded by the Scientific Research Foundation of Shantou University under Grant NTF24001T.

**Disclosure of Interests.** The authors declare that they have no known competing financial interests or personal relationships that could have appeared to influence the work reported in this paper.

## References

1. Chen, M., Lin, M., Li, K., Shen, Y., Wu, Y., Chao, F., Ji, R.: CF-ViT: A General Coarse-to-Fine Method for Vision Transformer. In: AAAI Conference on Artificial Intelligence. (2022)
2. Wei, Y., Zhao, L., Zheng, W., Zhu, Z., Zhou, J., Lu, J.: SurroundOcc: Multi-Camera 3D Occupancy Prediction for Autonomous Driving. In: 2023 IEEE/CVF International Conference on Computer Vision (ICCV), pp. 21672-21683. (2023)
3. Dan, J., Liu, Y., Xie, H., Deng, J., Xie, H., Xie, X., Sun, B.: TransFace: Calibrating Transformer Training for Face Recognition from a Data-Centric Perspective. In: 2023 IEEE/CVF International Conference on Computer Vision (ICCV), pp. 20585-20596. (2023)
4. Deng, H., Fang, Y., Huang, F.: Enhancing Adversarial Transferability on Vision Transformer by Permutation-Invariant Attacks. In: 2024 IEEE International Conference on Multimedia and Expo (ICME), pp. 1-6. (2024)
5. Huang, L., Gao, C., Zhuang, W., Liu, N.: Enhancing Adversarial Examples Via Self-Augmentation. In: 2021 IEEE International Conference on Multimedia and Expo (ICME), pp. 1-6. (2021)
6. Wan, C., Huang, F.: Adversarial Attack Based on Prediction-Correction. arXiv:2306.01809 (2023)
7. Long, Y., Zhang, Q., Zeng, B., Gao, L., Liu, X., Zhang, J., Song, J.: Frequency domain model augmentation for adversarial attack. In: Proc. Eur. Conf. on Computer Vision (ECCV), pp. 549-566. Springer (2022)
8. Wang, X., He, X., Wang, J., He, K.: Admix: Enhancing the transferability of adversarial attacks. In: Proc. IEEE Int. Conf. on Computer Vision (ICCV), pp. 16158-16167. (2021)
9. Wang, K., He, X., Wang, W., Wang, X.: Boosting adversarial transferability by block shuffle and rotation. In: Proc. IEEE Conf. on Computer Vision and Pattern Recognition (CVPR), pp. 24336-24346. (2024)

10. Li, Y., Bai, S., Zhou, Y., Xie, C., Zhang, Z., Yuille, A.L.: Learning Transferable Adversarial Examples via Ghost Networks. In: AAAI Conference on Artificial Intelligence. (2018)
11. Xie, C., Zhang, Z., Zhou, Y., Bai, S., Wang, J., Ren, Z., Yuille, A.L.: Improving transferability of adversarial examples with input diversity. In: Proc. IEEE Conf. on Computer Vision and Pattern Recognition (CVPR), pp. 2730-2739. (2019)
12. Dong, Y., Pang, T., Su, H., Zhu, J.: Evading defenses to transferable adversarial examples by translation-invariant attacks. In: Proc. IEEE Conf. on Computer Vision and Pattern Recognition (CVPR), pp. 4312-4321. (2019)
13. Lin, J., Song, C., He, K., Wang, L., Hopcroft, J.E.: Nesterov accelerated gradient and scale invariance for adversarial attacks. arXiv:1908.06281 (2019)
14. Dong, Y., Liao, F., Pang, T., Su, H., Zhu, J., Hu, X., Li, J.: Boosting adversarial attacks with momentum. In: Proc. IEEE Conf. on Computer Vision and Pattern Recognition (CVPR), pp. 9185-9193. (2018)
15. Saha, A., Subramanya, A., Patil, K., Pirsiavash, H.: Role of Spatial Context in Adversarial Robustness for Object Detection. In: 2020 IEEE/CVF Conference on Computer Vision and Pattern Recognition Workshops (CVPRW), pp. 3403-3412. (2019)
16. Tramèr, F., Kurakin, A., Papernot, N., Goodfellow, I., Boneh, D., McDaniel, P.: Ensemble adversarial training: Attacks and defenses. arXiv:1705.07204 (2017)
17. Liao, F., Liang, M., Dong, Y., Pang, T., Hu, X., Zhu, J.: Defense against adversarial attacks using high-level representation guided denoiser. In: Proc. IEEE Conf. on Computer Vision and Pattern Recognition (CVPR), pp. 1778-1787. (2018)
18. Xie, C., Wang, J., Zhang, Z., Ren, Z., Yuille, A.: Mitigating adversarial effects through randomization. arXiv:1711.01991 (2017)
19. Naseer, M., Khan, S., Hayat, M., Khan, F.S., Porikli, F.: A self-supervised approach for adversarial robustness. In: Proc. IEEE Conf. on Computer Vision and Pattern Recognition (CVPR), pp. 262-271. (2020)
20. Liu, Y., Chen, X., Liu, C., Song, D.X.: Delving into Transferable Adversarial Examples and Black-box Attacks. arXiv:1611.02770 (2016)
21. Szegedy, C., Vanhoucke, V., Ioffe, S., Shlens, J., Wojna, Z.: Rethinking the inception architecture for computer vision. In: Proc. IEEE Conf. on Computer Vision and Pattern Recognition (CVPR), pp. 2818-2826. (2016)
22. Szegedy, C., Ioffe, S., Vanhoucke, V., Alemi, A.: Inception-v4, inception-resnet and the impact of residual connections on learning. In: Proc. AAAI Conf. on Artificial Intelligence (AAAI). (2017)
23. He, K., Zhang, X., Ren, S., Sun, J.: Deep Residual Learning for Image Recognition. In: 2016 IEEE Conference on Computer Vision and Pattern Recognition (CVPR), pp. 770-778. (2015)
24. Long, Y., Zhang, Q., Zeng, B., Gao, L., Liu, X., Zhang, J., Song, J.: Frequency domain model augmentation for adversarial attack. In: Proc. Eur. Conf. on Computer Vision (ECCV), pp. 549-566. Springer (2022)